\documentclass[
]{ceurart}

\usepackage{listings}
\usepackage{algorithm}
\usepackage{algpseudocode}
\usepackage{amsmath}
\usepackage{tcolorbox}
\begin{document}

\copyrightyear{2024}
\copyrightclause{Copyright for this paper by its authors.
  Use permitted under Creative Commons License Attribution 4.0
  International (CC BY 4.0).}

\conference{AI4CC-IPS-RCRA-SPIRIT 2024: International Workshop on Artificial Intelligence for Climate Change, Italian Workshop on Planning and Scheduling, RCRA Workshop on Experimental evaluation of algorithms for solving problems with combinatorial explosion, and SPIRIT Workshop on Strategies, Prediction, Interaction, and Reasoning in Italy. November 25-28th, 2024, Bolzano, Italy \cite{Ai4ccIpsRcraSpirit2024}.}

\title{One to rule them all: natural language to bind communication, perception and action}

\author[1, 3]{Simone Colombani}[%
email=simone.colombani@studenti.unimi.it,
]

\author[2]{Dimitri Ognibene}[%
email=dimitri.ognibene@unimib.it,
]

\author[1]{Giuseppe Boccignone}[%
email=giuseppe.boccignone@unimi.it,
]

\address[2]{University of Milan, Italy}
\address[1]{University of Milano-Bicocca, Milan, Italy}
\address[3]{Oversonic Robotics, Carate Brianza, Italy}

\begin{abstract}
\emph{In recent years, research in the area of human-robot interaction has focused on developing robots capable of understanding complex human instructions and performing tasks in dynamic and diverse environments. These systems have a wide range of applications, from personal assistance to industrial robotics, emphasizing the importance of robots interacting flexibly, naturally and safely with humans.
\\
This paper presents an advanced architecture for robotic action planning that integrates communication, perception, and planning with Large Language Models (LLMs). Our system is designed to translate commands expressed in natural language into executable robot actions, incorporating environmental information and dynamically updating plans based on real-time feedback.
\\
The Planner Module is the core of the system where LLMs  embedded in a modified ReAct framework are employed to interpret and carry out user commands like \textit{`Go to the kitchen and pick up the blue bottle on the table'}. By leveraging their extensive pre-trained knowledge, LLMs can effectively process  user requests without the need to introduce new knowledge on the changing environment. The modified ReAct framework further enhances the execution space by providing real-time environmental perception and the outcomes of physical actions.
By combining robust and dynamic semantic map representations as graphs with control components and failure explanations, this architecture enhances a robot's adaptability, task execution efficiency, and seamless collaboration with human users in shared and dynamic environments.
Through the integration of continuous feedback loops with the environment the system can dynamically adjusts the plan to accommodate unexpected changes,  optimizing the robot's ability to perform tasks. Using a dataset of previous experience is possible to provide detailed feedback about the failure. Updating the LLMs context of the next iteration with suggestion on how to overcame the issue.
\\
This system has been implemented on RoBee, the cognitive humanoid robot developed by Oversonic Robotics, showcasing its adaptability and potential for integration across diverse environments. By leveraging LLMs and semantic mapping, the architecture enables RoBee to navigate and respond to real-time changes.}

\end{abstract}


\maketitle

\section{Introduction}
The integration of LLMs in robotic systems has opened new avenues for autonomous task planning and execution \cite{driess2023palm, ahn2022can}. These models demonstrate exceptional natural language understanding and commonsense reasoning capabilities, enhancing a robot’s ability to comprehend contexts and execute commands \cite{wang2024large, zeng2023large}. However LLMs are not be able to
plan autonomously, they need to be integrated in  architectures that enable them to understand the environment, the robot capabilities and state \cite{kambhampati2024llms}.
This research aims to empower robots to comprehend user requests and autonomously generate actionable plans in diverse environments.
\\
The efficacy of these plans relies on the robot’s understanding of its operating environment \cite{tellex2020robots}. To bridge this gap, our work employs scene graphs \cite{zhu2022scene} as a semantic mapping tool, offering a structured representation of spatial and semantic information within a scene. 
\\
In our approach, we leverage LLMs through in-context  \cite{dong2022survey}, which enables the models to learn and adapt based on the information provided in the context. Our work implements a modified version of the ReAct \cite{yao2023react} framework that expand  the context of LLMs with environmental information and execution feedback, allowing the model to plan and execute skills \cite{heuss2023concept} translating them into physical actions.

\paragraph{Motivation} 
The primary focus of our work is to enable robot to interact flexibly and robustly in dynamic and diverse environments with limited human intervention. Traditional robotic systems usually rely on static, pre-programmed instructions or closed world predefined knowledge and settings, limiting their adaptability to dynamic environments.
Interacting with humans in daily tasks within complex environments disrupts these assumptions. LLMs and VLM can provide open-domain knowledge to represent novel conditions without human intervention. 
However, these models are not  informed of the specific robot, task and settings at hand, that define what information can be relevant and necessary to find and reason about \cite{shanahan2006frame}. Exceeding in the level of detail may lead to impractical computational requirements and response time. Discarding crucial information, spatial or semantic, may lead to repeated failures due to the introduced  non-managed partial observability \cite{kaelbling1998planning}.  To find the relevant information may be too slow \cite{ognibene2014ecological}.
LLMs can still produce outputs that are logically inconsistent or impractical \cite{ji2023survey}, expecially if they are not integrated into systems that allow them to adapt to changes in the environment and the physical capabilities of the robots.
Finally task execution, robots may encounter unexpected situations, such as unanticipated obstacles, sensor errors, or changes in the environment that were not accounted for in the initial plan. Such scenarios necessitate robust error handling mechanisms and adaptive planning strategies that enable the system to reassess and modify its actions in real-time \cite{ruiz2022reasoning}.
By introducing execution controlling and failure management  into the planning process at different levels as well as retrieval of previous successful plans, we propose a solution to enhance the robustness and flexibility of LLM-based robotic systems. This approach ensures that the robot can effectively perceive changes in the environment and the failures that may arise from them, allowing it to adapt strategies in response to new challenges.

\paragraph{Proposed approach} Our system addresses the challenges of dynamic environments through a real-time perception module and a Planner module that integrates execution control, and failure management. It comprise a Controller that monitors the execution of tasks and detects errors, while the Explainer analyzes failures and suggests adjustments based on past experiences. This feedback loop enables adaptive re-planning, allowing the system to modify its actions as needed. Specifically, we propose the use of the ReAct \cite{yao2023react} framework, expanding its operational space with skills, physical actions of the robot and with perception action, to access information from the environment. By leveraging LLMs for natural language understanding and a perception system, the architecture supports autonomous task execution in dynamic scenarios.

\begin{figure}
    \centering
    \includegraphics[width=0.6\textwidth]{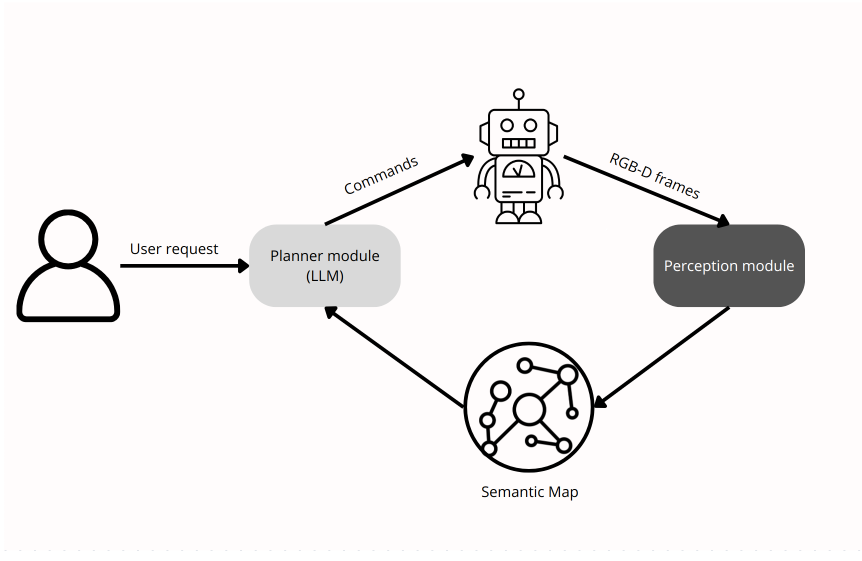}
    \caption{Architecture of the system.}
     \label{fig:system_architecture} 
\end{figure}

\section{Related works}
A substantial body of literature explores the utilization of LLMs for robotic task planning \cite{wang2024large, zeng2023large}.
\paragraph{LLM for robot planning}
Recent works highlight the potential of Large Language Models (LLMs) in robotic planning \cite{liang2023code, liu2023llm+, song2023llm}. DEPS \cite{wang2023describe} introduces an iterative planning approach for agents in open-world environments, such as Minecraft. It utilizes LLMs to analyze errors during execution and refine plans, improving both reasoning and goal selection processes. However, this approach has been primarily developed and tested in virtual environments, with notable differences in comparison to real-world settings due to the dynamic and unpredictable nature of physical environments. Additionally, DEPS does not leverage previous issues and solutions but relies solely on feedback from humans and vision-language models (VLMs).
\paragraph{Scene graph as environemental representation}
The use of scene graphs \cite{armeni20193d} as a means to represent the robot's environment has gained traction. \cite{liu2024delta} employs 3D scene graphs to represent environments and uses LLMs to generate Planning Domain Definition Language (PDDL) files. This method decomposes long-term goals into natural language instructions and enhances computational efficiency by addressing sub-goals. However, it lacks a mechanism for replanning based on feedback during execution, which could limit its adaptability in dynamic scenarios. SayPlan \cite{rana2023sayplan} integrates semantic search with scene graphs and path planning to aid robots in navigating complex environments through natural language. By combining these techniques, SayPlan simulates various scenarios to refine task sequences, which helps improve overall task performance in complex environments. However, it is reliance on static pre-built 3D scene graphs, hindering adaptability to dynamic real-world environments.
\paragraph{Replanning}
Replanning enables long-term autonomous task execution in robotics \cite{cashmore2019replanning}. DROC \cite{zha2024distilling} empowers robots to process natural language corrections and generalize that information to new tasks. It introduces a mechanism to distinguish between high-level and low-level errors, allowing more flexible plan corrections. However, DROC does not address the types of failures that may occur during plan execution, focusing instead on high-level corrections provided by users. 
\cite{skreta2024replan} supports autonomous long-term task execution by integrating LLMs for planning and VLMs for feedback. This approach adapts to changes in the environment through a structured component system that verifies and corrects plans as needed. Yet, the feedback is limited to what is visible to the robot’s camera, potentially overlooking other significant environmental changes.

\section{Architecture}
%
Our system is based on two components:
\begin{itemize}
    \item \textbf{Perception Module}: it is responsible for sensing and interpreting the environment. It builds and mantains a semantic map in the form of a directed graph that integrates both geometric and semantic information.
    
    \item \textbf{Planner Module}: it takes the information provided by the Perception Module to formulate plans and actions that allow the robot to perform specific tasks.
\end{itemize}

Figure \ref{fig:system_architecture} show how these components interact to allow the robot to understand its environment and act accordingly to satisfy user requests. The Perception module uses data provided by the robot's sensors to supply the semantic map to the Planner module, which in turn processes it to generate specific action plans. In what follows we precisely address the Planner Module while  details on the Perception Module will be provided in a separate article.
\\

\subsection{Planner module}
The architecture of the Planner module is designed to translate user requests, expressed in natural language, into specific actions executable by a robot. 
This module is responsible for understanding instructions, planning appropriate actions, and managing the execution of those actions in a dynamic environment.
The Planning module is composed by five sub-modules:
\begin{itemize}
    \item \textbf{Task Planner}: Translates user requests, expressed in natural language, into a sequence of high-level skills. 
    
    \item \textbf{Skill Planner}: Translates high-level skills into specific, low-level executable commands. 
    
    \item \textbf{Executor}: Executes the low-level actions generated by the Skill Planner. 
    
    \item \textbf{Controller}: Monitors the execution of actions and manages any errors or unexpected events during the process. 
    
    \item \textbf{Explainer}: Interprets the causes of execution failures by analyzing data received from the Controller and provides suggestions to the Task Planner on how to adjust the plan. 
\end{itemize}
\begin{figure}
    \centering
    \includegraphics[width=1.2\textwidth]{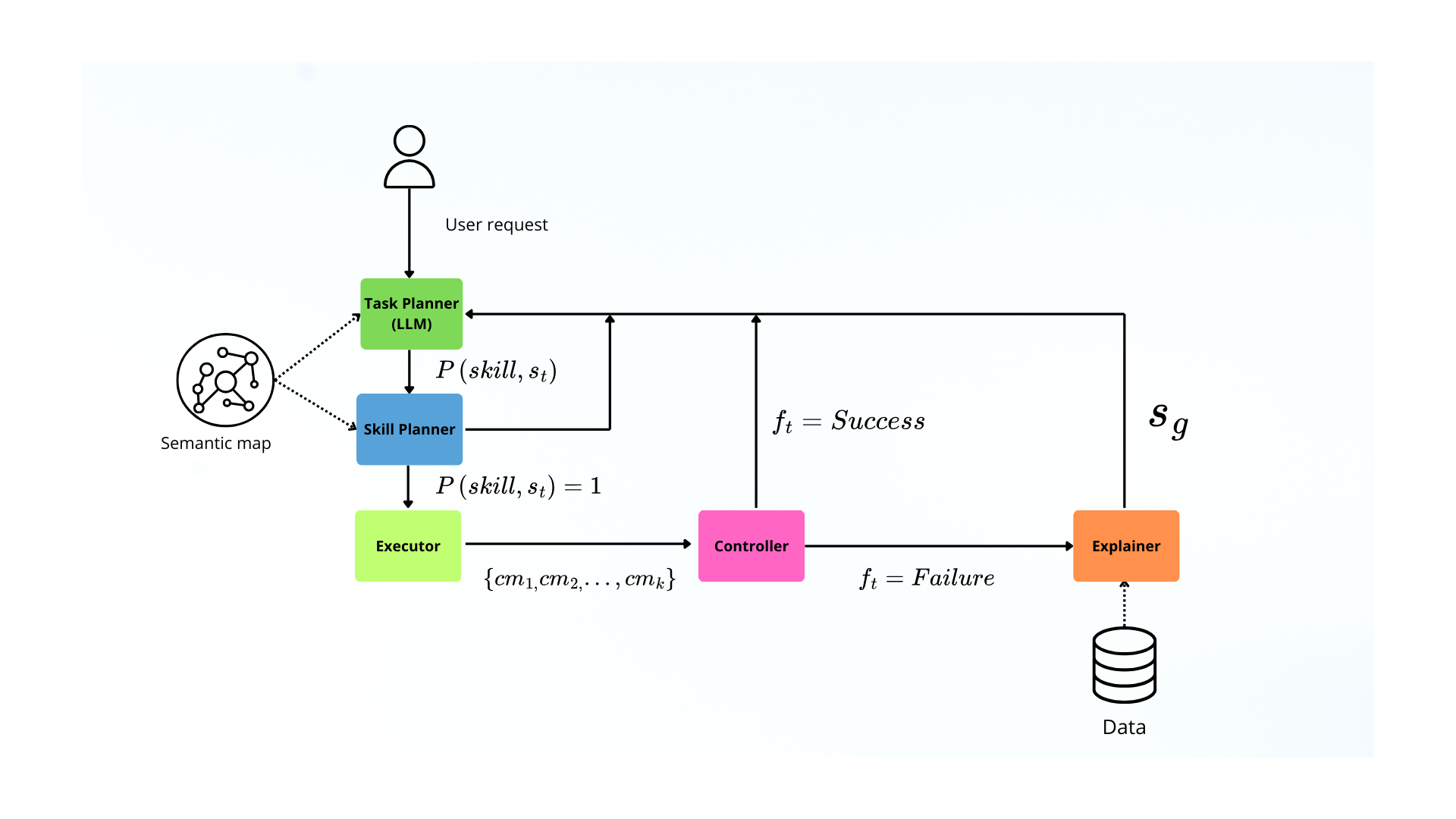}
    \caption{Architecture of the planner module.}
     \label{fig:architecture_planner} 
\end{figure}
The architecture of the planner module is shown in Figure \ref{fig:architecture_planner}. 
The main component of the system is the Task Planner, which receives the user's request and translates it into
a list of high-level "skills" that represent the robot's capabilities.
These skills include actions such as "PICK" (grasp an object), "PLACE" (place an object), and "GOTO" (move to a position). 
\\
\subsubsection{Task Planner}
The decision-making process of the Task Planner is driven by a \textbf{policy}, which is implemented as a LLM. A policy is a strategy or rule that defines how actions are selected based on the current state or context,\cite{geffner2014non}. 
\\
Task Planner is implemented using the ReAct framework \cite{yao2023react}, which alternates between reasoning and action phases during the process. In the reasoning phase, the Task Planner can access various \textbf{\textit{"perception"} actions} to gather information from the environment, such as the semantic map and the current state of the robot, and can execute one or more \textbf{\textit{"skill"} actions} to perform physical actions.
\\
The classical idea of ReAct is to augment the agent's action space to $\hat{A} = A \cup L$, where $L$ is the space of language-based reasoning actions. An action $\hat{a}_t \in L$, referred to as a \textit{"thought"} or reasoning trace, does not directly affect the external environment but instead updates the current context $c_{t+1} = (c_t, \hat{a}_t)$ by adding useful information to support future decision-making \cite{yao2023react}. In the classical idea there could be various types of useful thoughts, such as decomposing task goals and creating action plans, injecting commonsense knowledge relevant to task solving, extracting important parts from observations, tracking progress and transitioning action plans, handling exceptions and adjusting action plans, and so on, but always without modifying the physical environment, only embedding it within the context.
Interestingly, this approach mixes reasoning and action in a flexible manner. In the future, we will  analyse the potential of this approach also connecting to the planning-to-plan \cite{ognibene2009resources, ho2020people} and meta-reasoning \cite{russell1991principles,zilberstein1993anytime,  ackerman2017meta} concepts.
\\
In our work, we augment the agent's \cite{russell2016artificial} action space with two types of actions:
\begin{itemize}
   \item A \textit{skill} action $a_t \in A_{\text{skill}}$, which involves physically interacting with the environment, such as manipulating objects or navigating. The result of a skill action provides new feedback that updates the current context.
   \item A \textit{perception} action $a_t \in A_{\text{perception}}$, which involves accessing information from the environment, such as querying the semantic map or sensors, and integrating that information into the context.
\end{itemize}
The augmented action space is defined as:
\[
\hat{A} = A_{\text{skill}} \cup A_{\text{perception}} \cup L
\]
Thus, the LLM serves as the \textbf{policy} $\pi$ that selects different types of actions from the augmented action space and dynamically adapting the current context $c_t$ used to plan based on real-time information and reasoning.

\paragraph{Formal Description:}
The Task Planner's policy $\pi$, represented by the LLM, can be formalized as a function that maps the current context $c_t$ to an action $\hat{a}_t$ from the augmented action space $\hat{A}$:
\[
\pi: C \to \hat{A}, \quad \pi(c_t) = \hat{a}_t
\]
Where:
\begin{itemize}
    \item $C$ is the set of all possible contexts.
    \item $\hat{A}$ is the augmented action space $\hat{A} = A_{\text{skill}} \cup A_{\text{perception}} \cup L$.
    \item $c_t$ represents the current context at time $t$, which includes the state of the robot, the environment, and any past actions or thoughts.
    \item $\hat{a}_t \in \hat{A}$ is the action chosen by the policy, which can be a skill action $a_t \in A_{\text{skill}}$, a perception action $a_t \in A_{\text{perception}}$, or a reasoning trace $\hat{a}_t \in L$.
\end{itemize}

The context $c_t$ is updated based on the chosen action:
\begin{itemize}
    \item If $\hat{a}_t \in L$ (a reasoning action), the context updates to:
    \[
    c_{t+1} = (c_t, \hat{a}_t)
    \]
    This represents the thought process, where reasoning contributes new information without affecting the external environment.
  
    \item If $\hat{a}_t \in A_{\text{perception}}$ (a perception action), the result of querying the environment updates the context:
    \[
    c_{t+1} = (c_t,f_{\text{perception}} (\hat{a}_t))
    \]
    Here, $f_{\text{perception}}$ represents the function that gathers information and modifies the context based on the perception action's outcome.
  
    \item If $\hat{a}_t \in A_{\text{skill}}$ (a skill action), the robot interacts with the environment, and the context updates based on feedback from the physical action:
    \[
    c_{t+1} = (c_t, f_{\text{skill}}(\hat{a}_t))
    \]
    Where $f_{\text{skill}}$ is the function that captures the result of executing a physical skill, such as manipulating an object or moving to a location.
\end{itemize}

\subsubsection{Skill Planner} 
Once a high-level request for the execution of a skill is made, the Skill Planner is responsible for translating the high-level skills, provided by the Task Planner, into sequences of low-level commands executable by the robot. 
While the Task Planner focuses on understanding natural language and creating a general plan, the Skill Planner deals with the specific details of how each skill should be executed, considering the robot's state and the environment.
\\
Let a skill be represented in the following general form, defined by the Task Planner with specific syntax:
\[
\texttt{$SKILL\_NAME(param_1, param_2, \ldots, param_N)$}
\]
Where:
\begin{itemize}
    \item \texttt{SKILL\_NAME} is the name of the skill to be executed (e.g., \texttt{PICK}, \texttt{PLACE}, \texttt{GOTO}).
    \item \texttt{param\_1, param\_2, \ldots, param\_N} are parameters for the skill, such as the object to manipulate or the destination to navigate to.
\end{itemize}

Using a strict syntax ensures that the Skill Planner can correctly interpret the high-level commands without ambiguity. For instance, a natural language command like \emph{"Move near the table and grab the bottle"} would lack precision. The Skill Planner needs concrete parameters for the robot to act effectively.

\paragraph{Skill Planner workflow:}

The Skill Planner operates by performing three functions:

1. \textbf{Precondition Verification:}
   Before translating a skill into low-level commands, the Skill Planner verifies that the necessary preconditions for execution are met. Let $s_t$ represent the current state of the robot and the environment at time $t$, and $P(\text{skill}, s_t)$ denote a function for every skill that evaluates the preconditions for a given skill. The precondition check can be expressed as:
   \[
   P(\text{skill}, s_t) = 
   \begin{cases}
   1, & \text{if all preconditions are met} \\
   0, & \text{otherwise}
   \end{cases}
   \]
   For example, before executing the \texttt{PICK} skill, the following checks may be performed:
   \begin{itemize}
       \item The object is visible by the robot.
       \item The object is reachable for the robotic arm.
       \item The robotic arm is free.
   \end{itemize}
   If any of these conditions are not met ($P(\text{skill}, s_t) = 0$), the Skill Planner reports a failure to the Task Planner.

2. \textbf{Target nodes extraction:}
   Based on the parameters of the skill, the Skill Planner extracts the target nodes from the semantic map $\mathcal{M}$, which contains geometric and semantic information about the environment.
   Every node provides geometric information such as object's position and relevant context, which is then used to generate low-level commands.

3. \textbf{Generation of Low-Level Commands:}
When \( P(\text{skill}, s_t) = 1 \), the Skill Planner translate the skill into a sequence of low-level commands to control the robot behavior. In this system, we represent skill decomposition in commands as Hierarchical Task Networks (HTNs) that contains low-level commands executable by the robot.
Let $CM(\text{skill}, \text{node}, s_t)$ denote the function that translates the given skill into low-level commands  based on the target nodes extracted from the semantic map and current state. The output is a sequence of pre-modeled commands parameterized with the information of the robot state and the target nodes, $\{cm_1, cm_2, \ldots, cm_k\}$, where each command $cm_i$ directs specific components of the robot. 
Our implementation use HTNs solely on the breakdown of skills into commands without using them with advanced features like re-planning or error recovery of the commands. In this case, if any command fails, the entire skill fails, with no attempt at re-planning at the skill planner level.   
   The process can be represented as:
   
   \[
   \{cm_1, cm_2, \ldots, cm_k\} = CM(\text{skill}, \text{node}, s_t)
   \]

The Skill Planner is designed to be flexible and extendable. The skill functions $P$, and $CM$ can be adapted or extended to accommodate new skills, hardware, or environments.

\subsubsection{Executor} 
The Executor is responsible for directly interacting with the robot's hardware to execute the commands provided by the Skill Planner. 
It translates the low-level commands into physical actions by controlling various hardware elements such as motors, robotic arm grippers, and other actuators required for task execution.

Let the set of low-level commands generated by the Skill Planner be represented as above, i.e.,  $cm_1, cm_2, \ldots, cm_k\ = CM(\text{skill}, \text{node}, s_t)$, 
where $CM(\text{skill}, \text{node}, s_t)$ defines the sequence of commands based on the skill, the target node, and the current state of the robot and the environment.

The Executor is tasked with executing these commands on the physical robot. Let the state of the robot at time $t$ be denoted by $h_t$, and the function that maps a low-level command $c_i$ to an effect on the robot's state be denoted as $H(cm_i, h_t)$. The execution of a command at time $t$ can be described as:
\[
h_{t+1} = H(cm_i, h_t)
\]
where $h_{t+1}$ is the updated state after executing the command $cm_i$. This process is repeated for each command in the sequence $\{cm_1, cm_2, \ldots, cm_k\}$ until the entire skill is executed.

\paragraph{Executor workflow:}
\begin{itemize}
    \item \textbf{Command reception:} The Executor receives a set of low-level commands $\{cm_1, cm_2, \ldots, cm_k\}$ from the Skill Planner. Each command specifies a concrete action to be performed by the robot's hardware components.
    
    \item \textbf{Hardware interaction:} For each command $cm_i$, the Executor interacts with the robot’s hardware, adjusting the motors, grippers, and other actuators. This interaction can be represented by the function $H(cm_i, h_t)$ that determines the effect of a command on the robot's state $h_t$.

    \item \textbf{Command execution:} The Executor executes each command $cm_i$ in the sequence, ensuring that the robot's state transitions from state $h_t$ to $h_{t+1}$. Formally:
    \[
    h\_{t+1} = H(cm\_i, h\_t), \quad \forall i = 1, 2, \ldots, k
    \]
    After executing all commands, the robot's reaches the final state $h_{t+k}$, corresponding to the completion of the skill.
    
    \item \textbf{Real-Time feedback:} During execution, the robot's provides feedback on its current state. Let $f_{t}$ denote the feedback at time $t$, and $f_{t+1}$ be the updated feedback after executing command $cm_i$:
    \[
    f_{t+1} = F(cm_i, h_t)
    \]
    where $F$ is the feedback function. If unexpected feedback $f_{t+1}$ is received, the Executor can trigger adjustments to the plan or inform the Skill Planner of a potential issue.
\end{itemize}

Different robots may use different communication protocols, and hardware configurations. Therefore, the Executor must be adapted for each specific robot system, ensuring that it correctly interacts with the robot’s hardware.

\subsubsection{Controller} 
The Controller is responsible for monitoring the robot's status and the environment during command execution, ensuring that they are carried out as planned. After each command is executed, the Executor sends feedback indicating either success or failure. If a failure occurs, it results in the failure of the entire skill. Upon the completion of all commands, a success feedback will indicate the successful execution of the skill.
\\
Denote $f_t$ the feedback from the Executor at time $t$. The Controller processes $f_t$ to determine the outcome of the executed skills. The feedback can be classified into two categories: success and failure.

\paragraph{Feedback processing:}
\begin{itemize}
    \item \textbf{Success:} If the feedback $f_t$ indicates successful execution of a command and it is the last command to execute, the skill is considered successfully completed, the Controller sends a positive acknowledgment to the Task Planner to continue the planning process. However, if the feedback indicates success but the command is not the last one, the Controller waits for the execution of the next command:
    \[
    \text{if } f_t = \text{Success} \implies \text{Task Planner continues}
    \]

    \item \textbf{Failure:} If a failure occurs during the execution of any command, the planned skill fails and the Controller generates a failure message $m_f$ that includes the reason for the failure. This message is sent to the Explainer. Let $e_t$ represent the specific error detected at time $t$. The failure message can be represented as:
    \[
    m_f = \text{Failure}(e_t)
    \]
    where $e_t$ can include various error reasons such as obstacles detected, non-executable trajectories, or environmental changes. 
\end{itemize}

The Controller's operation is highly dependent on the specific robot system in use, as it relies on the characteristics of the robot and the employed software system. In a ROS environment, for example, the Controller interacts with ROS nodes that control the robot's hardware.
In our work, RoBee, described in section \ref{sec:robot}, has a system that allows to obtain feedback on the execution of commands.

\subsubsection{Explainer} 
The Explainer component plays a critical role in enhancing the planning process by providing insights to the Task Planner when failures occur during the execution phase. 
After receiving the failure reason, the Explainer searches a dataset $\mathcal{D}$ for previous instances of similar failures. This dataset comprises records of failures associated with specific skills and user requests. Let $\mathcal{D}_{r_f}$ denote the subset of the dataset containing records of failures and solutions related to the same skill and error message. The dataset has been manually built based on previous experiences, desired behaviors, and expected failures.

The search can be expressed as:
\[
\mathcal{D}_{r_f} = \{ (s_k, u_r, r_f) \in \mathcal{D} \,|\, s_k = \text{skill\_name}, \, e_r = r_f, \, u_r \sim \text{user\_request} \}
\]

where:
\begin{itemize}
    \item $s_k$ is the skill being executed (e.g., PICK).
    \item $u_r$ represents the specific user request associated with the failure.
    \item $r_f$ is the failure reason provided by the Controller
     \item $u_r \sim \text{user\_request}$ indicates that the user request in the dataset is similar to the current user request.
\end{itemize}
Rather than searching for an exact match to the user's request, the Explainer assesses the similarity of the user's request ($u_r$) to the instances in the dataset linked to the suggestion, using cosine similarity in our approach \cite{rahutomo2012semantic}. This method enables the system to identify the most relevant past instances, even when the user's requests are not identical.
\\
Once relevant instances are identified, the Explainer analyzes these cases to generate a suggestion $s_g$ for the Task Planner on how to proceed. The suggestion is structured as follows:
\[
s_g = \text{Suggest}(\mathcal{D}_{r_f})
\]
\\
For instance, if the Controller reports the failure reason:
\[
r_f = \text{"Cannot execute the approach movement for the PICK skill, object too far"}
\]
The Explainer analyzes this failure and may find a previous instance where the robot successfully resolved a similar issue. It could recommend a command to the Task Planner:
\[
s_g = \text{"Use the GOTO skill to move near the object to pick"}
\]
This suggestion enables the Task Planner to adjust its strategy effectively, moving the robot closer to the object before attempting the PICK action again.
\\
The suggestions provided by the Explainer can be tailored to accommodate specific behaviors of the robot. This adaptability can be achieved by modifying the parameters of the data utilized to generate the suggestions. Thus, the Explainer enhances the resilience of the system, facilitating its ability to adapt to changing conditions and recover from errors efficiently.

\section{Behavior example of the system}
To illustrate the proposed system’s behavior in a dynamic environment, consider a scenario where the robot is tasked with picking up a bottle from a table in a room and placing it in a designated area in other room. In this example, the system must react to unexpected changes, such as the bottle being moved by an external agent during the task.

\paragraph{Example}
\textbf{User Request:} The user provides the instruction: "Go to the table in the kitchen, pick up the bottle, and place it on the table in the bedroom."
\textbf{Perception Module:} The robot senses the environment and generateds a semantic map, which includes the pose and types of objects.
\textbf{Task Planner:} The request is interpreted, and the Task Planner begins the planning procedure. It starts by calling perception actions to gather information about the environment and the robot's state. Using the results of these perception actions, the Task Planner determines the next steps in its execution by calling skill actions and incorporating the outcomes of each skill into the planning process. LLM interprets the request and reasons as follows.

\begin{tcolorbox}[width=1\textwidth, colback=white, colframe=gray!50]
\textbf{User Request:} "Go to the table in the kitchen, pick up the bottle, and place it on the table in the bedroom."  \\
\textbf{Robot State:} The robot is currently in the bedroom and has both the right and left arms empty.\\ [1ex]
\textbf{Thought:} The task is to fetch the bottle from the kitchen and place it on the table in the bedroom. I need to know all the rooms where the robot can go to. \\
        \textbf{Perception action:} \texttt{GetMapRooms()} \\ 
        \textbf{Observation:} The robot identifies the rooms: \texttt{[kitchen, bedroom]}. \\[1ex]
        \textbf{Thought:} Now, I will navigate to the kitchen to find the bottle. \\
        \textbf{Skill action:} \texttt{GOTO(kitchen)} \\ 
        \textbf{Observation:} The robot moves to the kitchen. \\[1ex]
        \textbf{Thought:} I should identify the specific location of the bottle within the kitchen. \\
        \textbf{Perception action:} \texttt{GetObjectInRoom(kitchen)} \\ 
        \textbf{Observation:} The robot finds the following objects in the kitchen: \texttt{[sink on the table\_1, stove on the table\_1, bottle, table\_1, table\_2]}. \\[1ex]
        \textbf{Thought:} I will now move to the location of the bottle. \\
        \textbf{Skill action:} \texttt{GOTO(bottle)} \\ 
        \textbf{Observation:} The robot moves to the bottle's location in the kitchen. \\[1ex]
         \textbf{Thought:} I will pick up the bottle now. \\
        \textbf{Skill action:} \texttt{PICK(bottle, right)} \\ 
        \textbf{Observation:} The robot can't see the bottle to pick, find a bottle in the semantic map and try again to pick it. \\[1ex]
        \textbf{Thought:} The bottle is not visible where expected, I need to check the location of the bottle again. \\
        \textbf{Perception action:} \texttt{GetObjectInRoom(kitchen)} \\ 
        \textbf{Observation:} The robot finds the following objects in the kitchen: \texttt{[sink on the table\_1, stove on the table\_1, bottle on the table\_2, table\_1, table\_2 ]}. \\[1ex]
        \textbf{Thought:} I will go to the table to pick up the bottle instead. \\
        \textbf{Skill action:} \texttt{GOTO(table\_2)} \\ 
        \textbf{Observation:} The robot moves to the table. \\[1ex]
        \textbf{Thought:} I will pick up the bottle from the table now. \\
        \textbf{Skill action:} \texttt{PICK(bottle, right)} \\ 
        \textbf{Observation:} The robot successfully picks up the bottle. \\[1ex]
        \textbf{Thought:} Next, I need to navigate to the bedroom to place the bottle. \\
        \textbf{Skill action:} \texttt{GOTO(bedroom)} \\ 
        \textbf{Observation:} The robot moves to the bedroom. \\[1ex]
        \textbf{Thought:} I need to locate the table in the bedroom where I will place the bottle. \\
        \textbf{Perception action:} \texttt{GetObjectInRoom(bedroom)} \\ 
        \textbf{Observation:} The robot sees: \texttt{[table, bed, lamp on the table]} in the bedroom. \\[1ex]
        \textbf{Thought:} I will approach the table to place the bottle. \\
        \textbf{Skill action:} \texttt{GOTO(table)} \\ 
        \textbf{Observation:} The robot approaches the table. \\[1ex]
        \textbf{Thought:} Finally, I will place the bottle on the table. \\
        \textbf{Skill action:} \texttt{PLACE(bottle, right)} \\ 
        \textbf{Observation:} The robot places the bottle on the table successfully.
\end{tcolorbox}

\textbf{Skill Planner:} For each high-level skill, once the preconditions are met, the Skill Planner translates the choosen skill in a sequence of low-level commands, such as motor movements for navigation, arm articulation for picking, and placing actions. For example once the \texttt{PICK(bottle, right)} skill is planned, it can be translated and divided into the following phases with relative commands:
\begin{itemize}
    \item Approach: The robot arm moves towards the object's position, making any necessary adjustments to align correctly, and opens the gripper.
    \item Grasp: The robot activates the gripping mechanisms to seize the object. This phase includes closing the gripper and verifying the grasp.
    \item Lifting: The robot lifts the object from the surface it is on.
\end{itemize}

\textbf{Execution}: The Executor begins executing the planned skill, which is composed of a sequence of commands by the Skill Planner. The Executor follows the ordered steps to achieve the goal.
For example with the skill \texttt{PICK(bottle, right)}, the Executor receive the list of command and execute:
\begin{itemize} 
\item Execute approach: The robot arm moves towards the object's position and open the gripper.
\item Execute grasp: This phase includes closing the gripper and verifying the grasp.
\item Execute lifting: The robot lifts the object from the surface it is on.
\end{itemize}
Thus, when an unexpected event occurs, such as the bottle being moved or is not reachable the executor may raise a failure message.

\textbf{Controller} and \textbf{Explainer} interaction:
\begin{itemize}
    \item The Controller detects that the object is no longer in the expected location and sends a failure message to the Explainer.
    \item The Explainer analyzes the failure, referencing previous instances where objects were moved unexpectedly. It suggests the Task Planner to re analyse the semantic map and update the object’s location.
\end{itemize}
\textbf{Re-planning:} Based on the suggestion, the Task Planner issues a new plan:
\begin{itemize}
    \item Execute \texttt{GOTO(table)} to go near the identified bottle.
    \item After locating the bottle on the table, the robot updates its actions and proceeds to execute the remaining tasks.
\end{itemize}

This example demonstrates how the system adapts in real-time, allowing for continuous task execution even in dynamic and unpredictable environments.

\paragraph{Planning algorithm}
We now formalize this process in the form of an adaptive planning algorithm. In this algorithm, the used LLM is a generalist model such as \textit{Llama 3 70B Instruct} \cite{dubeyllama}, whose behavior we influence through in-context learning \cite{dong2022survey}.
\begin{algorithm}[H] 
    \caption{Planning with extedend ReAct Framework}
    \label{alg}
    \begin{algorithmic}[1]
        \State \textbf{Input:} User request $r$, Robot state $R_{s}$ 
        \State \textbf{Output:} Execution of user request

        \Procedure{Planning}{$r, M$} 
        \State $C_0 \gets \text{InitializeLLMContext}(r,\ M, \ R_{s})$ 
        \While{not goal achieved}            
            \State $action \gets \text{TaskPlanner}(r, C_0)$ \Comment{Get first skill}
            \If{$action = \text{\textit{"Skill"}}$}            
                \State $commands \gets \text{SkillPlanner}(skill, C_t)$ \Comment{Translate skill into low-level commands}
                \State $success \gets \text{Executor}(commands)$ \Comment{Execute commands}
                
                \If{$success = \text{False}$}
                    \State $failureMsg \gets \text{Controller}(C_t)$ \Comment{Detect failure}
                    \State $c_{t} \gets \text{Explainer}(failureMsg)$ \Comment{Generate suggestion}
                \Else
                    \State $c_{t} \gets \text{Skill succesfully executed}$
                \EndIf
            \Else
                \State $c_{t} \gets \text{CallPerceptionAction}()$ \Comment{Reading semantic map from Perception Module}
            \EndIf
                \State $C_{t+1} \gets \text{UpdateContext}(C_t)$ \Comment{Update context}
                \State $skill \gets \text{TaskPlanner}(r, C_{t+1})$ \Comment{Get next skill based on updated context}
            \EndWhile
        \EndProcedure 
    \end{algorithmic} 
\end{algorithm}

This algorithm shows the adaptive behavior of the system by incorporating feedback loops that facilitate real-time re-planning. By alternating between action and reasoning phases, the robot can continuously adapt to changes, ensuring task success even in unpredictable environments.

\section{Robot Hardware}
\label{sec:robot}
\begin{figure}
    \centering
    \includegraphics[width=0.6\textwidth]{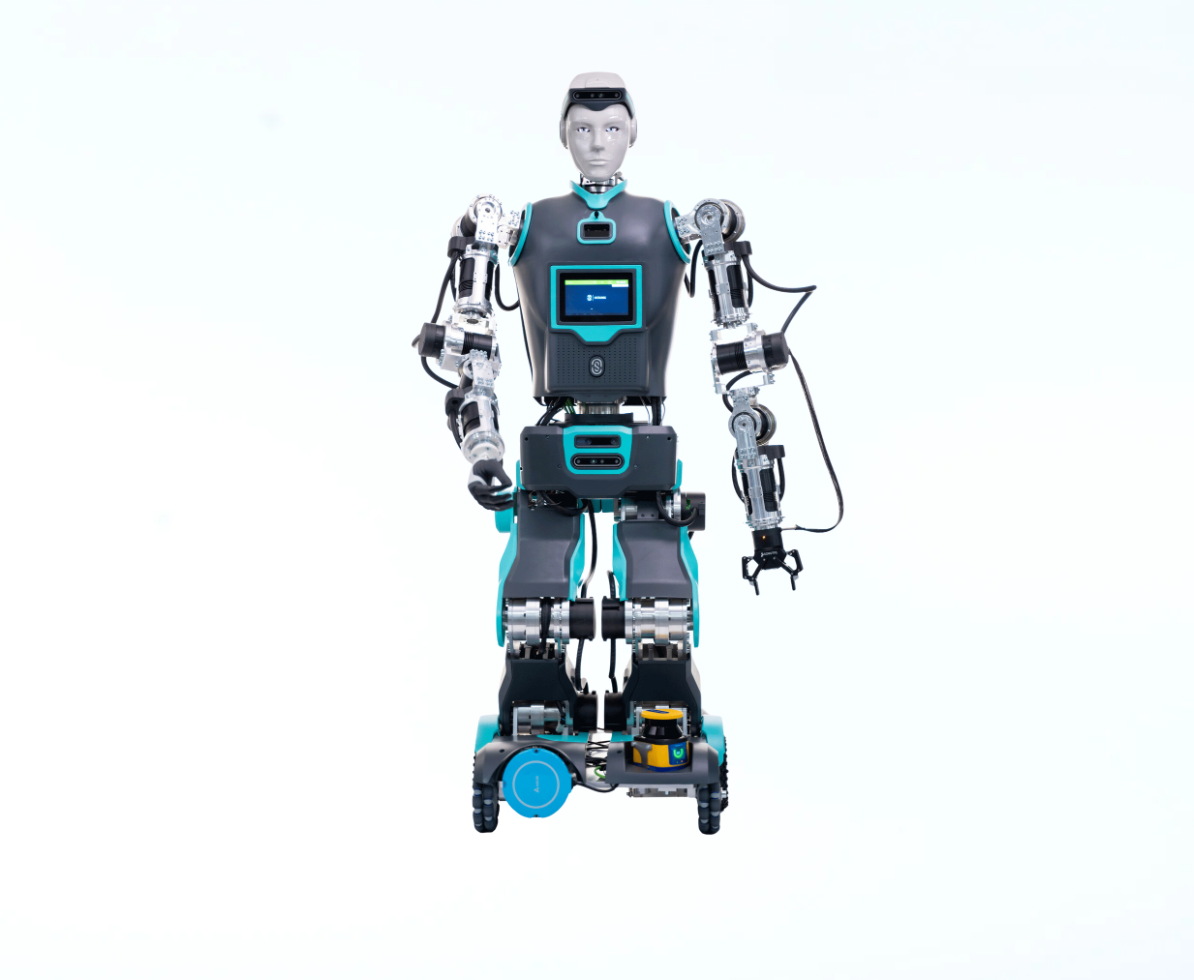}
    \caption{Robee, humaniod robot developed by Oversonic Robotics.}
     \label{fig:robee} 
\end{figure}
The system was implemented using RoBee, a cognitive humanoid robot developed by Oversonic Robotics. RoBee measures 160 cm in height and weighs 60 kg. It has 32 degrees of freedom, enabling highly flexible movement. The robot is equipped with multiple sensors, including cameras, microphones, and force sensors. 
\\
The cameras provide real-time visual data, supporting navigation and object recognition tasks. The microphones facilitate audio input, enabling speech recognition and interaction through natural language processing. The force sensors are used for handling objects, allowing RoBee to adjust grip force based on the characteristics of the item being manipulated, enhancing precision and safety during interactions.
\\
RoBee's mechanical structure includes two arms capable of bimanual manipulation, each capable of handling objects weighing up to 5 kg. The system includes a torso and leg system designed for balance and mobility. RoBee is equipped with LIDAR sensors for real-time environment mapping and obstacle detection. These LIDAR sensors enable the robot to navigate autonomously through complex environments, ensuring safe operation in shared spaces. The combination of autonomous navigation technologies and LIDAR-based detection enhances the ability of RoBee to move efficiently and avoid collisions in dynamic industrial environments.
\\
In addition to its physical capabilities, RoBee integrates with cloud-based systems, allowing for remote monitoring, task scheduling, and data analytics. 
\\
The Planner-module takes into account RoBee’s embodiment, ensuring that the system is aligned with the robot's capabilities such as its degrees of freedom, sensor suite, and ability to perform manipulation and navigation.

\section{Preliminary results}
\begin{figure*}
    \centering
    \begin{minipage}{0.33\textwidth}
        \includegraphics[width=\textwidth]{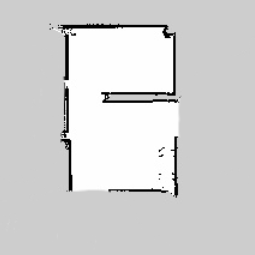}
    \end{minipage}\hfill
    \begin{minipage}{0.33\textwidth}
        \centering
    \includegraphics[width=\textwidth]{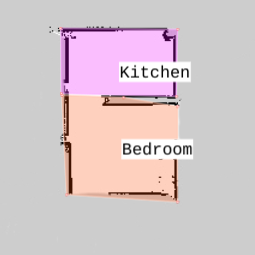}
    \end{minipage}
     \begin{minipage}{0.33\textwidth}
        \centering
    \includegraphics[width=\textwidth]{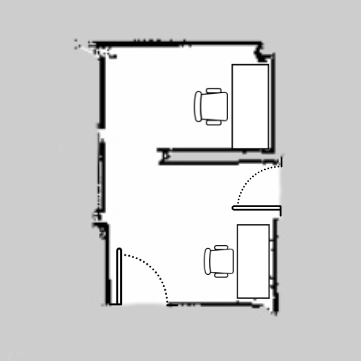}
    \end{minipage}
    \label{fig:test_maps}
    \caption{Environment used during the execution of experiments.}
\end{figure*}

Preliminary experiments were conducted in a simulated environment replicating two main rooms: a \textit{kitchen} and a \textit{bedroom}, as illustrated in Figure \ref{fig:test_maps}. 

During the experiments, three types of requests were tested, each varying in complexity:

\begin{itemize}
    \item \textbf{Simple requests}: direct commands that involve only one skill. For example, \textit{"Pick up the bottle in front of you"}, where the task planner needs only to identify the parameters and activate the appropriate skill.

    \item \textbf{Moderately complex requests}: tasks that require the robot to perform multiple skills in sequence, as explicitly described in the command. An example is \textit{"Go to the kitchen, pick up the bottle, and bring it to the table in the bedroom"}, which involves multiple skills. These tasks require a higher level of complexity, with planning across several steps and handling potential failures.

   \item \textbf{Complex requests}: such as \textit{"I'm thirsty, can you help me?"}, which were more open-ended and required the robot to interpret the task and break it down into multiple steps.
\end{itemize}

The results in table \ref{tab:planner_results} showed that the system performed well with simple requests, followed by moderately complex ones. However, the success rate for complex requests was significantly lower, with only 25\% of the tasks completed correctly. This lower performance was attributed to the system's difficulty in understanding and managing ambiguous or under-specified instructions.
\\
It is important to note that these are preliminary results, and further analysis is ongoing. A thorough evaluation of the data is currently underway, including a comparison with the state of the art in robot task execution and natural language understanding. This will allow for a deeper understanding of the system's strengths and areas for improvement.

\begin{table}[h]
    \centering
    \begin{tabular}{|c|c|c|}
        \hline
        \textbf{Request type} & \textbf{Number of attempts} & \textbf{Success rate} \\
        \hline
        \textit{Simple requests} & 30 & 90\% \\
        \hline
        \textit{Moderately complex requests} & 20 & 75\% \\
        \hline
        \textit{Complex requests} & 10 & 25\% \\
        \hline
    \end{tabular}
    \caption{Number of attempts and success rate for each request type}
    \label{tab:planner_results}
\end{table}

\section{Conclusions}
The proposed planning system exhibits notable strengths, particularly its adaptability and seamless with the robot's diverse  set of skill for executing complex tasks. The system's core advantage lies in its ability to interpret user commands through natural language processing, converting them into high-level actions that are further refined into low-level, executable tasks. By integrating real-time environmental feedback from the Perception Module through an extended version of ReAct framework, the system can dynamically adjust to unexpected situations, such as obstacles or execution failures. This adaptability is supported by an architecture, where the Task Planner, Skill Planner, Controller, and Explainer components work in harmony to ensure smooth task execution even in changing environments.
\\
One of the system's key strengths is its ability to manage error recovery through feedback loops, allowing the robot to adapt quickly to failures during task execution. The Explainer module provides on the fly suggestions to modify the plan based on past errors, enhancing the system’s validity. The use of semantic maps and scene graphs provides the robot with a structured understanding of its environment, ensuring that actions are contextually accurate and responsive to real-world conditions.
\\
The integration of  LLMs, perceptual feedback, and flexible task planning mechanisms makes the system highly versatile for complex, dynamic environments. Its implementation on RoBee, the humanoid robot developed by Oversonic Robotics, has demonstrated its practical potential, positioning it as a valuable tool for applications requiring advanced human-robot interaction and adaptability in unpredictable settings.
\\
In the future, other than extending the low level skill set available, we will investigate the possibility to autonomously expand the Explainer dataset as well as providing similar information directly to the Task Planner,  increasing  flexibility and reliability and reducing the number of  re-planning events. We will also study capability of the system to proactively acquire information about the environment \cite{ognibene2014ecological} and human partners both through sensors \cite{ognibene2013towards} and communication strategies, leveraging the potential for proactive information gathering behaviours of LLMs \cite{patania-etal-2024-large,ren2023robots,magnini-2024-toward}. Moreover, it will be crucial to assess the reliability of the system both at the planning level as well as the communication level, considering the introduction of embodiment and environment while the limitation in pragmatic understanding of LLM are still to be understood \cite{magnini-2024-toward,martinenghi-etal-2024-von,martinenghi2024llms}.

\acknowledgments
Special thanks to Oversonic Robotics for enabling the implementation of this project using their humanoid robot, RoBee.

\bibliography{bibliography}

\section{Online Resources}
More information about RoBee and Oversonic Robotics are available:
\begin{itemize}
\item \href{https://oversonicrobotics.com/robee-humanoid-robot/?lang=en}{RoBee},
\item \href{https://oversonicrobotics.com/?lang=en}{Oversonic Robotics}
\end{itemize}
\end{document}